\documentclass[sigconf, nonacm]{acmart}
\AtBeginDocument{%
  }

\usepackage{hyperref}
\usepackage{url}
\usepackage{booktabs}       % professional-quality tables
\usepackage{amsmath,amsfonts}
\usepackage{nicefrac}       % compact symbols for 1/2, etc.
\usepackage{microtype}      % microtypography
\usepackage{xcolor}         % colors
\usepackage{graphicx}
\usepackage{booktabs}
\usepackage{bbm}
\usepackage{multirow}
\usepackage{subcaption}
\usepackage{algpseudocode}
\usepackage{bbm}
\usepackage{dsfont}
\usepackage{tikz}
\usepackage{pgfplots}
\usepackage{amsthm}
\usepackage{xspace} 
\usepackage{tabularray}
\UseTblrLibrary{booktabs}
\usepackage{makecell} 
\usepackage{pifont}
\newcommand{\xmark}{\ding{55}}

\newcommand{\ours}{mpLLM\xspace}
\setcopyright{none}
\renewcommand\footnotetextcopyrightpermission[1]{} 

\begin{document}

\title{Multimodal LLM With Hierarchical Mixture-of-Experts for VQA on 3D Brain MRI}

\author{Arvind Murari Vepa}
\affiliation{%
  \institution{UCLA}
  \department{Computer Science}
  \country{USA}
}
\email{amvepa@ucla.edu}
% add department remove location (try to save space)
% can share affiliation
% emphasize they are part of radiology
% fabien, emphasize department of medicine
\author{Yannan Yu}
\affiliation{%
  \institution{UCSF}
  \department{Radiology}
  \country{USA}
  }
\email{yannan.yu@ucsf.edu}

\author{Jingru Gan}
\affiliation{%
  \institution{UCLA}
  \department{Computer Science}
  \country{USA}
}
\email{jrgan@cs.ucla.edu}

\author{Anthony Cuturrufo}
\affiliation{%
  \institution{UCLA}
  \department{Computer Science}
  \country{USA}
}
\email{acc@cs.ucla.edu}

\author{Michael F. Romano}
\affiliation{%
  \institution{UCSF}
  \department{Radiology}
  \country{USA}
}
  \email{michael.romano@ucsf.edu
  }

\author{Weikai Li}
\affiliation{%
  \institution{UCLA}
  \department{Computer Science}
  \country{USA}
}
\email{weikaili@cs.ucla.edu}

\author{Fabien Scalzo}
\affiliation{%
  \institution{UCLA}
  \department{Medicine}
  \country{USA}
}
\email{fab@cs.ucla.edu}

\author{Wei Wang}
\affiliation{%
  \institution{UCLA}
  \department{Computer Science}
  \country{USA}
}
\email{weiwang@cs.ucla.edu}

\author{Yizhou Sun}
\affiliation{%
  \institution{UCLA}
  \department{Computer Science}
  \country{USA}
}
\email{yzsun@cs.ucla.edu}

\renewcommand{\shortauthors}{Vepa et al.}

\begin{abstract}
Multiparametric 3D brain MRI (mpMRI) is central to neuroradiology, but producing tumor location, appearance, size, and involvement of critical structures for neurosurgical planning remains challenging. We introduce \ours, a multimodal LLM for visual question answering (VQA) on mpMRI that produces clinically interpretable tumor descriptors (e.g., volume, morphology, extent, and coarse localization) as an adjunct to clinical expertise for referring neurosurgeons. \ours uses a prompt-conditioned hierarchical mixture-of-experts (MoE) to fuse multiple 3D sequences via routing over modality- and token-level projection experts, enabling data-efficient end-to-end training without large-scale image–report pretraining. To address limited paired image–text supervision, we propose a synthetic VQA protocol that derives clinically grounded questions and answers from expert segmentation annotations and is validated with radiologist collaboration. Across multiple mpMRI datasets, \ours improves over strong medical VLM baselines by +5.5 points on average (+9.1\% relative) and increases radiologist-rated clinical acceptability by +15.9 points (+46.6\% relative). Our study features three main contributions: (1) the first VQA dataset for 3D brain mpMRI, (2) a hierarchical MoE architecture for joint reasoning over interrelated 3D sequences, and (3) expert-supported evidence of clinical utility. Source code is available at https://github.com/arvindmvepa/mpllm, and we will release the dataset upon publication.
\end{abstract}

\begin{CCSXML}
<ccs2012>
   <concept>
       <concept_id>10010405.10010444.10010449</concept_id>
       <concept_desc>Applied computing~Health informatics</concept_desc>
       <concept_significance>500</concept_significance>
       </concept>
   <concept>
       <concept_id>10010147.10010257.10010293.10010294</concept_id>
       <concept_desc>Computing methodologies~Neural networks</concept_desc>
       <concept_significance>500</concept_significance>
       </concept>
   <concept>
       <concept_id>10010147.10010178.10010179.10010182</concept_id>
       <concept_desc>Computing methodologies~Natural language generation</concept_desc>
       <concept_significance>500</concept_significance>
       </concept>
   <concept>
       <concept_id>10010147.10010178.10010224</concept_id>
       <concept_desc>Computing methodologies~Computer vision</concept_desc>
       <concept_significance>500</concept_significance>
       </concept>
 </ccs2012>
\end{CCSXML}

\ccsdesc[500]{Applied computing~Health informatics}
\ccsdesc[500]{Computing methodologies~Neural networks}
\ccsdesc[500]{Computing methodologies~Natural language generation}
\ccsdesc[500]{Computing methodologies~Computer vision}

\keywords{Multimodal LLM, Mixture-of-Experts, VQA, Synthetic Data, Medical Imaging}
\begin{teaserfigure}
  \includegraphics[width=\textwidth]{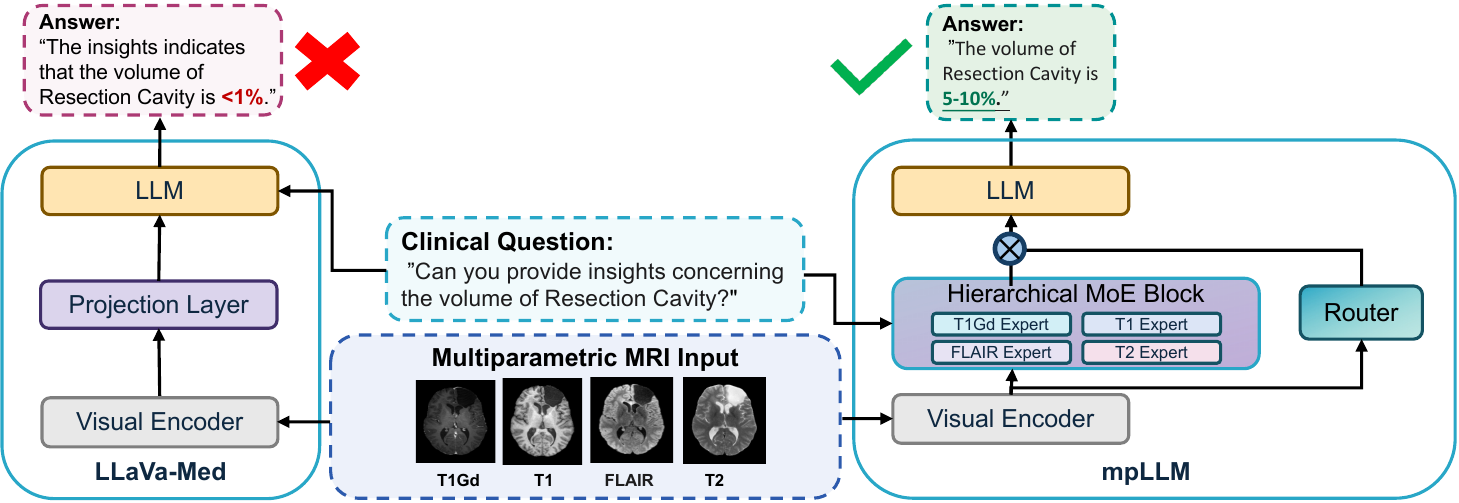}
  \caption{High-level comparison between LLaVA-Med and \ours. While LLaVA-Med uses a standard projection layer, our method uses a hierarchical MoE block which ingests both the prompt and imaging to produce prompt-conditioned vision tokens that leverage all the 3D modalities. }
  \Description{High-level comparison between LLaVA-Med and \ours. While LLaVA-Med uses a standard projection layer, our method uses a hierarchical MoE block which ingests both the prompt and imaging to produce prompt-conditioned vision tokens that leverage all the 3D modalities. }
  \label{fig:intro}
\end{teaserfigure}


\maketitle

\section{Introduction}
\label{sec:intro}

Multiparametric MRI (mpMRI) plays a significant role in diagnosing, grading, treating, and assessing treatment responses for brain tumors and other intracranial lesions \citep{sawlani2020multiparametric, wang2022multi, cherubini2016importance}. Producing clear, consistent descriptions of tumor size, morphology, extent, and localization is time-consuming, yet critical for downstream decision-making. As a result, a growing body of work has developed recognition and localization models to support clinical interpretation \citep{ghadimi2025deep, rathore2018deriving, wang2022multi, li2023multi, osman2019multi}.

A key unmet need is interactive, queryable assistance: neurosurgical planning and counseling depend not only on detecting tumors but on communicating clinically interpretable descriptors—where the lesion is, what it looks like, how large it is, and their involvement of critical structures. However, existing models have limited clinical utility because clinicians cannot effectively pose natural language queries about mpMRI. While 3D vision-language models (VLMs) have been developed for other imaging domains, current architectures do not naturally leverage the interdependencies among mpMRI modalities \citep{li2023llava, wu2023towards, bai2024m3d, xin2025med3dvlm}. Additionally, the standard multi-image approach multiplies the number of vision tokens by the number of images, which significantly increases computational constraints \citep{wu2023towards}.

We introduce \ours, a prompt-conditioned hierarchical mixture-of-experts (MoE) for VQA over mpMRI. Our approach is an extension of the LLaVa architecture \cite{liu2023visual}. \ours replaces the standard single projection with a prompt-conditioned MoE projection that routes across modality-level and token-level projection experts to fuse multiple 3D sequences into an efficient representation for the language model. Unlike approaches that require training separate modality-specific encoders, we use lightweight projection functions that train end-to-end with the language model during fine-tuning. 

To address limited paired image–text supervision in neuroimaging, we pair \ours with a synthetic VQA protocol that derives clinically grounded question–answer pairs from expert segmentation annotations, developed in collaboration with radiologists to ensure clinical relevance. In contrast to prior work, we fine-tune directly on the resulting VQA supervision via next-token prediction without pretraining on paired image–report data, and we train an end-to-end multi-task head to improve task proficiency and support reliable evaluation. In summary, our research makes these key contributions:

\begin{enumerate}
\item In collaboration with radiologists, we introduce a synthetic VQA protocol that produces the first VQA dataset for 3D brain mpMRI.
\item We design \ours, a multimodal LLM with prompt-conditioned hierarchical MoE routing that leverages interdependencies among 3D mpMRI modalities.
\item We provide empirical evidence of clinical utility through expert-supported validation and clinically relevant downstream evaluation.
\end{enumerate}

\begingroup
\setlength{\textfloatsep}{8pt plus 2pt minus 2pt} % space below top floats
\begin{figure*}[t]
    \centering
    \includegraphics[width=.75\linewidth,trim=6pt 6pt 6pt 6pt]{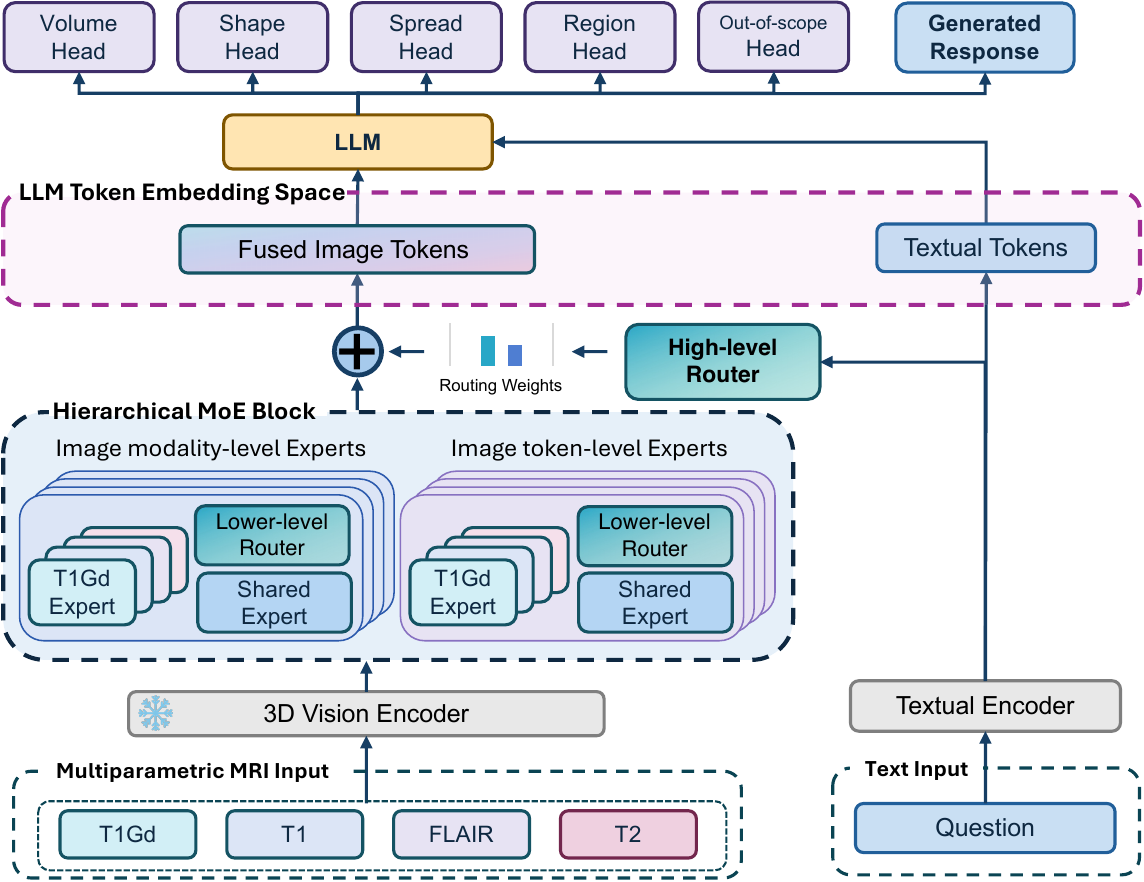}
    \caption{Detailed overview of our \ours pipeline.}
    \Description{Detailed overview of our \ours pipeline.}
    \label{fig:framework}
\end{figure*}
\endgroup

\section{Related Work}
\label{sec:related_work}

\paragraph{Medical vision-language models}
Most vision-based medical multimodal LLMs can be broadly classified into CLIP-based discriminative models \citep{radford2021learning, wang2022medclip, eslami2023pubmedclip, zhang2023biomedclip, xu2024whole, zhou2024knowledge, huang2023visual} and LLM decoder-based generative models \citep{zhang2023pmc, li2023llava, moor2023med}. Although discriminative models have proven helpful for various image recognition tasks, they possess limited utility in generation tasks such as VQA or report generation. Several popular generative models including MedVInt \citep{zhang2023pmc}, LLaVA-Med \citep{li2023llava}, and MedFlamingo \citep{moor2023med} share very similar architectures. However, these architectures and many others \citep{liu2024pefomed, lin2023pmc, li2023self, zhu2024mmedpo, lin2025healthgpt, zhang2025medunifier, nath2024vila, guo2025din} are designed specifically for 2D medical imaging and are not tailored to handle multiple 3D medical image modalities.

Although several 3D VLMs exist for natural images 
\citep{zhu2024unifying, li2024uni3dl, zhu20233d}, they require access to extremely large annotated datasets, which are often unavailable in medical contexts. While a few 3D VLMs have been developed for medical imaging, these methods have certain limitations. In one recent paper, researchers adapted the LLaVA-Med architecture to utilize spatial pooling and pretrain a 3D vision encoder with 700k radiology images \citep{bai2024m3d}. In another paper, researchers pretrain segmentation modules to generate brain imaging reports \citep{lei2024autorg}. In recent work, researchers exploit vision-language pretraining for CT report generation \citep{liu2023t3d, chen2024medblip, blankemeier2024merlin, xin2025med3dvlm, cao2025boosting, li2025towards}. However, these prior works assume a large paired imaging-report pretraining dataset, which is infeasible to collect and imposes a significant training burden. 

Furthermore, previous methods focus on report generation instead of VQA, leading to less precise feedback regarding model strengths and weaknesses. Additionally, some models train directly on segmentation annotations \citep{lei2024autorg, rui2024brainmvp}, which are impractical to obtain, especially for novel use cases. Moreover, none of the previously discussed methods are tailored to handle multiple interdependent 3D image modalities, like in mpMRI, as input.

\paragraph{Mixture-of-experts}

Previous work in MoE has concentrated on training and inference efficiency \citep{shazeer2017outrageously, lepikhin2020gshard, fedus2022switch, zoph2022st, liu2024deepseek}, transfer learning \citep{li2022sparse, zhong2022meta}, class imbalance \citep{han2024node}, and multi-domain information \citep{zhang2024m3oe}. There have also been earlier efforts with multimodal LLMs, covering sparsity learning \citep{lin2024moe}, task interference \citep{shen2025mome}, and embedding models \citep{li2024your}. MoE also has various applications in the medical field. These applications include addressing missing modalities \citep{yun2025flex, novosad2024task, liu2024mixture}, fairness \citep{wang2025fair}, pediatric care \citep{HuyTa_PedCLIP_MICCAI2025}, parameter reduction and efficiency \citep{jiang2024med, nathani2024knowledge}, super resolution \citep{lin2021generalised}, and segmentation of multimodal medical imaging \citep{zhang2025foundation, jiang2024m4oe}. 

Related to our work, several studies have employed MoE with VLMs to select between vision encoders and vision-language projections \citep{li2025cumo, zong2024mova, wang2023image, ma20253d}. In CuMo \citep{li2025cumo}, the authors incorporate MoE in the MLP layers in the vision encoder as well as projection function; however, they do not utilize prompt-conditioning, do not focus on combining multiple inputs, and require a complex training scheme (our approach requires a single stage of instruction fine-tuning). In MoVA \citep{zong2024mova}, the authors utilize MoE with task-specific vision encoders. In practice, training vision encoders is challenging and resource intensive in the medical space \citep{rui2024brainmvp}. Improving the LLaVa projection function with a MoE block uses far less parameters and can be trained end-to-end with the LLM using next-token prediction. The previous approach also does not focus on combining multiple inputs. Finally, in 3D-MoE \citep{ma20253d}, the authors utilize MoE within the LLM and do not focus on combining multiple inputs. Additionally, no existing research has explored using MoE to project multiple interrelated 3D image modalities into the language embedding space.

\paragraph{Medical VQA Datasets}

One of the primary challenges in report generation is evaluation: lexical metrics such as BLEU, ROUGE‑L, and BERTScore have been shown to correlate poorly with radiologist evaluations \citep{yu2023evaluating}. In contrast, VQA allows for more granular and interpretable model evaluation. While there are several medical VQA datasets, many focus on 2D imaging \citep{liu2024gemex, liu2021slake, he2020pathvqa, lau2018dataset}. In a prior work, researchers used a scene graph generator to generate surgical VQA \citep{yuan2024advancing}. In a recent work, researchers extracted multi-task questions from structured lung cancer screening data \citep{niu2025medical}. While there are existing VQA datasets for 2D brain MRI \cite{safari2025performance, shehzad2025brain, peng2025omnibrainbench}, there are no existing VQA datasets for 3D brain mpMRI due to a significant lack of source data for VQA extraction. In our work, we remedy this by collaborating with radiologists to generate clinically relevant synthetic data from expert annotated segmentation annotations and leverage ChatGPT-4o for dataset augmentation. Other prior works have also leveraged commercial LLMs to generate VQA data from expert annotations and study metadata in the medical setting \citep{zhang2023pmc, gautam2024kvasir, butsanets2025radimagenet}. To contextualize dataset reliability and demonstrate practical value, we report data quality metrics, radiologist acceptability ratings of model-generated responses, and performance on clinically relevant downstream tasks.

\section{Methodology}

Brain mpMRI has several 3D imaging modalities. For a given modality $m$, let $I_m \in \mathbb{R}^{C \times D \times H \times W}$ denote the corresponding 3D volume, where $C$, $D$, $H$, and $W$ represent the channel, depth, height, and width, respectively. A 3D vision encoder $h_{\text{vis}}$ maps each modality to a sequence of image tokens
$v_m = h_{\text{vis}}(I_m) \in \mathbb{R}^{T \times d_{V}}$, where $T$ is the number of image tokens and $d_{V}$ is the vision embedding dimension. We apply spatial pooling to reduce the token length (reusing the symbol $T$ for the pooled length for simplicity) and then concatenate the pooled tokens across the $M$ image modalities before passing them to the hierarchical MoE.  

Let $v \in \mathbb{R}^{M \times T \times d_{V}}$ denote the concatenated image modality embeddings. The hierarchical MoE projects these embeddings into the LLM space,
$e = \mathrm{MoE}(v, t) \in \mathbb{R}^{M \times T \times d_{T}}$, where $\mathrm{MoE}$ denotes the hierarchical MoE block, $t$ represents the text prompt, and $d_{T}$ is the LLM embedding dimension. In practice we flatten the modality and token dimensions and provide $(t, e)$ as a soft prompt to the LLM for multi-task prediction and text generation. A detailed visualization of our approach can be seen in Figure~\ref{fig:framework}.

\subsection{Hierarchical mixture-of-experts for multiparametric MRI projection}

In what follows, we use the term \emph{expert} exclusively for projection modules that map vision features into the LLM embedding space, and the term \emph{router} for the MLPs that output mixing weights over experts. Each high-level \emph{expert block} therefore consists of a router together with its associated projection experts.

\paragraph{High-level router} Our hierarchical MoE architecture includes a high-level router $r^{(h)}$ that assigns weights over a set of high-level expert blocks $\{\mathcal{E}^{(h)}_{1}, \dots, \mathcal{E}^{(h)}_{H}\}$, where $H$ is the number of high-level experts. These expert blocks operate at the image modality and image token levels. The router is implemented as a two-layer MLP that takes as input the final hidden state of the language model corresponding to the text prompt $t$. It produces a normalized weight distribution over expert blocks:
$\pi^{(h)}(t) = \mathrm{softmax}(r^{(h)}(t)) \in \mathbb{R}^{H}$. Since task information is embedded within the text prompt, the router implicitly infers the task, enabling high-level experts to specialize in different task proficiencies.

\paragraph{High-level image modality-level and image token-level experts} Our hierarchical MoE includes high-level experts operating at different granularity levels: image modality-level and image token-level. Each high-level expert block consists of a two-layer MLP low-level router $r^{(l)}$ and an associated set of low-level projection experts $\{W^{(l)}_{1}, \dots, W^{(l)}_{L}\}$, where $L$ is the number of low-level experts within the block.

The image modality-level expert block uses a low-level router that takes as input the concatenated [CLS] tokens from all image modalities (e.g., T1, T2) and outputs modality-level weights over the corresponding low-level experts: $\pi^{(l)}_{\text{mod}}(v) = \sigma(r^{(l)}_{\text{mod}}(v)) \in \mathbb{R}^{L}$. In contrast, the token-level expert block uses a low-level router that receives, for each token position $i$, the $i$-th image tokens from all modalities and outputs token-level weights $\pi^{(l)}_{\text{tok}}(v) = \sigma(r^{(l)}_{\text{tok}}(v)) \in \mathbb{R}^{L \times T}$. As discussed in prior work \citep{li2024hierarchical}, providing weights at different granularities improves task performance by enhancing domain generalizability.

\paragraph{Low-level image modality-specific and image modality-agnostic (shared) experts} Each low-level expert $W$ represents a projection transformation from the vision encoder embedding space to the LLM embedding space: $W: R^{N_{I} \times d_{I}} \rightarrow R^{N_{I} \times d_{T}}$. We utilized a simple linear transformation for the projection transformation as in the original LLaVa paper \citep{liu2023visual}. Each image modality embedding is processed through a modality-specific expert and a modality-agnostic (shared)  expert. The modality-specific expert emphasizes extracting image modality-specific features, whereas the modality-agnostic expert focuses on deriving common features from all image modalities. The parameters for the modality-specific expert are unique to each image modality (T1Gd, T1, T2, and FLAIR), while those for the modality-agnostic expert are consistent across all image modalities.

Each image modality is passed through both low-level experts and then summed embedding dimension-wise. The overall formulation for the hierarchical MoE is as follows:

\begin{equation}
\label{eq:moe}
\begin{aligned}
\mathrm{MoE}(v, t)
&= \sum_{h=1}^{H} \pi^{(h)}_{h}(t) \, \cdot \\
&\quad \sum_{m=1}^{M}
    \Big(
      \alpha^{(h)}_{m}\, W^{(h)}_{m}(v_{m})
      + \beta^{(h)}_{m}\, W^{(h)}_{\text{shared}}(v_{m})
    \Big)
\end{aligned}
\end{equation}

where $h$ indexes the high-level expert blocks, $m$ indexes the image modalities, and $\alpha^{(h)}_{m}$ and $\beta^{(h)}_{m}$ are the modality-specific and modality-agnostic weights produced by the corresponding low-level router (which sum to one within an expert block $h$). Our validation experiments on the GLI dataset found that the optimal number of high-level experts was 16, corresponding to the number of labels times the number of tasks. %We summed the features from the different experts instead of concatenating them to reduce the number of image tokens, which significantly enhances training and inference efficiency and, according to our validation tests, had no adverse effect on performance. More discussion related to MoE hyperparameters is in the ablation study (section \ref{sec:abl}) and the Appendix.

The fused image token embeddings are combined with the text prompt token embeddings. Then, these embeddings are input into the LLM decoder at the token embedding layer for multi-task prediction and text generation.

\begin{table*}
\setlength{\textfloatsep}{8pt plus 2pt minus 2pt}
\centering
\caption{Comparison of task performance for all models on all datasets with accuracy metric with standard deviation.}
\SetTblrInner{rowsep=0.95pt}
\resizebox{\linewidth}{!}{
\begin{tblr}{colspec = {cccccccc}}
    \toprule
\makecell{Dataset} & \makecell{Method} & \makecell{Volume} & \makecell{Region} & \makecell{Shape} & \makecell{Spread} & \makecell{Mean} \\
\midrule
\SetCell[r=6]{c} GLI 
& RadFM \citep{wu2023towards} & 12.9$\pm$0.8 & 68.2$\pm$0.5 & 17.9$\pm$0.9 & 17.0$\pm$0.9 & 29.0$\pm$0.4 \\
& Med3DVLM \citep{xin2025med3dvlm} & 29.7$\pm$1.1 & 72.9$\pm$0.5 & 39.5$\pm$1.2 & 35.3$\pm$1.2 & 44.3$\pm$0.6 \\
& LLaVA-Med \citep{li2023llava}  & 39.1$\pm$1.3  & 73.9$\pm$0.5 & 53.9$\pm$1.2 & 50.8$\pm$1.2 & 54.4$\pm$0.7 \\
& Merlin \citep{blankemeier2024merlin} & 55.1$\pm$1.2 & 75.9$\pm$0.5 & 52.1$\pm$1.2 & 49.2$\pm$1.2 & 58.1$\pm$0.6 \\
& M3D \citep{bai2024m3d} & 39.7$\pm$1.3 & 73.9$\pm$0.5 & 54.1$\pm$1.1 & 52.0$\pm$1.2 & 54.9$\pm$0.7 \\
& \ours (Ours) & \textbf{62.5}$\pm$\textbf{1.2} & \textbf{83.0}$\pm$\textbf{0.4} & \textbf{59.2}$\pm$\textbf{1.2} & \textbf{56.5}$\pm$\textbf{1.2} & \textbf{65.3}$\pm$\textbf{0.6} \\
\midrule
\SetCell[r=6]{c} MET 
& RadFM & 15.8$\pm$1.4 & 69.9$\pm$0.9 & 13.0$\pm$1.4 & 12.2$\pm$1.3 & 27.7$\pm$0.7 \\
& Med3DVLM & 41.5$\pm$1.9 & 70.2$\pm$0.9 & 32.7$\pm$1.9 & 33.2$\pm$1.8 & 44.4$\pm$1.0 \\
& LLaVA-Med & 47.0$\pm$2.1 & 68.6$\pm$1.0 & 38.3$\pm$2.0 & 35.2$\pm$1.9 & 47.3$\pm$1.0 \\
& Merlin & 58.7$\pm$2.0 & 72.8$\pm$1.0 & 48.2$\pm$1.9 & 43.0$\pm$2.0 & 55.7$\pm$1.1 \\
& M3D  & \textbf{65.7}$\pm$\textbf{1.9} & 73.6$\pm$0.9 & 48.2$\pm$2.0 & 42.2$\pm$2.0 & 57.4$\pm$1.0 \\
& \ours (Ours) &  \textbf{65.7}$\pm$1.9 & \textbf{76.6}$\pm$\textbf{0.9} & \textbf{51.3}$\pm$\textbf{1.9} & \textbf{53.0}$\pm$\textbf{2.0} & \textbf{61.6}$\pm$\textbf{1.0} \\
\midrule
\SetCell[r=6]{c} GoAT 
& RadFM & 13.0$\pm$1.0 & 64.3$\pm$0.6 & 34.6$\pm$1.4 & 27.6$\pm$1.2 & 34.9$\pm$0.6 \\
& Med3DVLM & 31.2$\pm$1.4 & 65.1$\pm$0.6 & 55.5$\pm$1.5 & 48.3$\pm$1.4 & 50.0$\pm$0.7 \\
& LLaVA-Med & 59.3$\pm$1.4 & 76.7$\pm$0.5 & \textbf{76.3}$\pm$\textbf{1.2} & 66.6$\pm$1.3 & 69.7$\pm$0.6\\
& Merlin & 56.3$\pm$1.5 & 75.1$\pm$0.6 & 68.9$\pm$1.4 & 51.5$\pm$1.5 & 63.0$\pm$0.7 \\
& M3D & 56.9$\pm$1.4 & 75.3$\pm$0.5 & 75.5$\pm$1.3 & 65.4$\pm$1.4 & 68.3$\pm$0.7 \\
& \ours (Ours) & \textbf{63.0}$\pm$\textbf{1.4} & \textbf{77.4}$\pm$\textbf{0.5} & 73.4$\pm$1.3 & \textbf{66.9}$\pm$\textbf{1.4} & \textbf{70.2}$\pm$\textbf{0.7} \\
\midrule
\end{tblr}
}
\label{tab:results}
\end{table*}

\subsection{Training objectives}

\subsubsection{Synthetic VQA protocol}
\label{sec:auto_vqa}

Because of the lack of brain mpMRI VQA data, we propose a novel method of synthetic VQA generation that leverages the expert annotated brain mpMRI segmentation data. We perform the following steps: 1) compute continuous quantities from masks, 2) map to quantities to report-style discrete labels, and 3) insert discrete labels into paraphrased QA templates. 

To generate relevant VQA data, we consult with radiologists to identify important topics that can be extracted from the label masks, focusing on mask volume relative to brain volume \citep{kaifi2023review}, brain region localization \citep{lau2018dataset}, shape \citep{ismail2018shape}, and spread \citep{islam2019brain}. For each label mask, we compute the quantities using standard formulas and validate the thresholds with synthetic masks and a subset of data. To emulate the subjectivity found in medical reports, we categorize each of the quantities based on their magnitude using terminology similar to that found in medical reports. We employ a rules-based method to assign medical terms to the quantities. We assign ``N/A'' if the label is not found.

\paragraph{Volume} To calculate the relative mask volume, we determine the number of mask pixels and divide by the number of brain pixels in the volume (which are the nonzero pixels in the skull-stripped  T1 image modality). The subjective labels we use are  ``$<1\%$'', ``$1-5\%$'', ``$5-10\%$'', ``$10-25\%$'', ``$25-50\%$'', and ``$50-75$\%''.

\paragraph{Region} For the BraTS GLI volumes, we use the Nibabel python library \citep{abraham2014machine} to register the volumes to the AAL atlas (version SPM12) \cite{rolls2020automated} and extract the following brain regions: ``frontal'', ``parietal'', ``occipital'', ``temporal'', ``limbic'', ``insula'', ``subcortical'', and ``cerebellum''. For the BraTS MET and GoAT volumes, we register the volumes to the LPBA40 atlas (in SRI24 space) \citep{shattuck2008construction} and extract the following brain regions: ``frontal'', ``parietal'', ``occipital'', ``temporal'', ``limbic'', ``insula'', ``subcortical'', ``cerebellum'', and ``brainstem''. The percent coverages of the segmentation masks with the atlases are 67.3\%, 70.9\%, and 57.7\% for the GLI, MET, and GoAT datasets respectively.

\paragraph{Shape} We first quantify each mask's overall size and compute classical 3‑D shape metrics (sphericity, elongation, flatness, solidity, compactness). If the mask is tiny, it is classified as ``focus''; otherwise, we classify it as ``round,'' ``oval,'' ``elongated,'' or ``irregular'' by comparing its sphericity and elongation values to empirically chosen thresholds that correspond to near‑sphere, mildly flattened, and strongly stretched geometries.

\paragraph{Spread} We identify all disconnected islands, noting the largest as the ``core,'' and compute what proportion of the total mask volume it occupies. If there is only one island, the pattern is ``single lesion''; if multiple islands are present but the core retains $\geq$85\% of the volume, it is described as ``core with satellite lesions''; otherwise, when no dominant island exists, the distribution is marked ``scattered lesions.''

\paragraph{Question-answer pair generation} After computing the previous quantities for each label mask, we create a dataset that simulates the natural variability of human input. First, we consider all combinations of the four major tasks to create multi-task question-answer pairs. After we have the 15 question-answer pair types, we use ChatGPT-4o to generate approximately 3000 perturbations of each question-answer pair (without affecting the label and answer term) that emulates the language a radiologist would use. We also add question-answer pairs with partially out-of-scope and completely out-of-scope tasks to improve the model’s self-awareness of its capabilities. Thus, for each label and mpMRI in each dataset, we sample four multitask question-answer pairs without replacement such that each major task is addressed in at least one question-answer pair, one partially out-of-scope question-answer pair, and one completely out-of-scope question-answer pair. Examples of generated question-answer pairs can be seen in the Appendix in Table~\ref{tab:qual_vqa}. The answers are used as supervision for next-token prediction for the multimodal LLM.

\subsubsection{Multi-task heads} For increased task proficiency and more accurate task evaluation, we train a multi-task head end-to-end with the multimodal LLM. After providing the soft-prompt to the multimodal LLM, we extract the hidden state from the last layer and apply task-specific heads (which consist of a single linear layer) to generate multi-task predictions. For volume, shape, spread, and out-of-scope task identification, the task is multi-class classification, and the associated loss is categorical cross-entropy; whereas for region localization, the task is multi-label classification, and the associated loss is multi-label binary cross-entropy. These losses are added to the next-token prediction loss to produce our multi-task loss:

\begin{equation}
\label{eq:loss}
\begin{aligned}
\mathcal{L}
&= \mathcal{L}_\textrm{Next-token}
 + \mathcal{L}_\textrm{Volume}
 + \mathcal{L}_\textrm{Region} \\
&\quad
 + \mathcal{L}_\textrm{Shape}
 + \mathcal{L}_\textrm{Spread}
 + \mathcal{L}_\textrm{Out-of-scope}
\end{aligned}
\end{equation}

\section{Experiments}

\subsection{Datasets details}
\label{sec:dataset_details}

For our synthetic VQA protocol, we leverage the 2024 Brain Tumor Segmentation (BraTS) challenge \citep{labella2024braintumorsegmentationbrats}, which provides a standardized benchmarking environment for automated brain tumor segmentation. All datasets comprise of co-registered multiparametric MRI scans (T1, T1Gd, T2, FLAIR) at 1mm$^3$ resolution, skull-stripped and manually annotated by experts.  To enable fair comparison and manage GPU memory, all BraTS sequences were resampled to $32 \times 256 \times 256$. This allowed for compatibility with baseline methods, such as M3D \citep{bai2024m3d} and Med3DVLM \citep{xin2025med3dvlm}. We consider three challenges in BraTS: GLI, MET and GoAT. The challenges are collected from over ten institutions and encompass diverse pathological contexts and imaging protocols.

\paragraph{GLI (Adult Glioma Post Treatment)} Dataset focuses on post-treatment diffuse glioma segmentation and consists of multi-institutional routine post-treatment clinically-acquired multiparametric mpMRI scans of glioma. The task requires the delineation of enhancing tumor (ET), non-enhancing tumor core (NETC), surrounding FLAIR hyperintensity (SNFH), and resection cavity (RC) \citep{de20242024}.

\paragraph{MET (Brain Metastases)} Dataset contains a retrospective compilation of treatment-naive brain metastases mpMRI scans obtained from various institutions under standard clinical conditions. The challenge addresses the segmentation of small metastatic lesions using a 3-label system (NETC, SNFH, ET) and demonstrates variable tumor component distribution across cases \citep{moawad2024brain}.

\paragraph{GoAT (Generalizability Across Tumors)} Dataset assesses algorithmic generalizability across different tumor types (i.e., different number of lesions per scan, lesion sizes, and locations in the brain), institutions (i.e., different MRI scanners, acquisition protocols), and demographics (i.e., different age, sex, etc.). The challenge uses consistent labels (necrosis, edema/invaded tissue, and enhancing tumor) despite varying tumor morphology to evaluate algorithm adaptability to new disease types with limited training data \citep{de20242024, moawad2024brain, labella2023asnr, kazerooni2024brain, adewole2023brain}.

To generate the train, validation, and test sets, we randomly sample 80\%, 10\%, and 10\% from the imaging studies. For GLI we generated 31,104, 4,176, and 3,624 question-answer pairs for the train, validation, and test sets based on 1,621 mpMRIs. For MET, we generated 9,090, 1,368, and 1,260 question-answer pairs for the train, validation, and test sets, based on 651 mpMRIs. For GoAT, we generated 19,440, 2,430, and 2,448 question-answer pairs for the train, validation, and test sets, based on 1351 mpMRIs. All the generated questions pass a rules-based filter which checks for proper formatting. Additionally, all the generated test set questions pass an automated data cleaning step using ChatGPT-4o to ensure high quality evaluation.

\paragraph{Clinical validation} We collaborated with two radiologists who annotated 20 mpMRIs from the BraTS-GLI test set, 10 mpMRIs from the BraTS-MET test set, and 10 mpMRIs from the BraTS-GoAT test set with questions spanning four tasks and four findings for the BraTS-GLI dataset and three findings for the BraTS-MET and BraTS-GoAT datasets, yielding 560 questions. We used 10 annotated mpMRIs from BraTS-GLI for validation to improve the task label thresholds for the synthetic data. We used the other 30 annotated mpMRIs to evaluate the inter-annotator agreement between the synthetic data and each radiologist, the model predictions and each radiologist, and both radiologists (Table \ref{tab:clin_dataset_val} and \ref{tab:clin_dataset_model_val}). 

%We observe that in general the task accuracies are very similar, indicating that the synthetic data quality is comparable to radiologist-annotated data. We note a relatively lower accuracy for the region task, which is dependent on the quality of the brain-atlas registration when the imaging datasets were originally created. Despite the region label having more noise, we found that including the region label improved model performance on other tasks. 

To assess the quality of our synthetic questions, a radiologist evaluated the clarity of 160 synthetic questions from 10 mpMRIs from the BraTS-GLI test set using a binary scoring system (1 = valid, 0 = invalid). Prior to the LLM automated data cleaning, the synthetic questions achieved a 92.2\% validity rate. After automated data cleaning, the synthetic questions achieved a 95.9\% validity rate. More details about the datasets can be found in Appendix~\ref{sec:more_dataset_details} and ~\ref{sec:addl_ablation}.

\subsection{Experimental settings}
\label{sec:exp_settings}

\paragraph{Models} In our experiments, we utilized the Phi-3-Mini-4K-Instruct LLM. We also explored utilizing the Llama models and chose Phi-3 because of the increased efficiency and negligible performance benefits of the Llama models. For versatility and generality, we utilize the 3D Vision Transformer (3D ViT) \citep{dosovitskiy2020image} as the vision encoder and use medically pretrained weights \citep{bai2024m3d}.

\paragraph{Training} We fine-tune the multimodal LLM using the loss defined in Equation~\ref{eq:loss} on the VQA training dataset. We freeze the vision encoder while unfreezing the hierarchical MoE and LLM weights. We train the model on the train dataset for 2 epochs. The LLM is trained with LoRA, setting $r$ to 16 and $alpha$ to 32, with a dropout of 0.1. We employ a cosine learning rate scheduler that starts at a learning rate of $2.0\times 10^{-4}$.

\paragraph{Baseline models} We compare our approach to several baseline models, including LLaVA-Med \citep{li2023llava}, M3D \citep{bai2024m3d}, Merlin \citep{blankemeier2024merlin}, Med3DVLM \citep{xin2025med3dvlm}, and RadFM \citep{wu2023towards}. To process the multiple 3D MRI image modalities, we use a multi-image approach, in which we concatenate the image tokens generated from each MRI image modality from a shared projection layer and vision encoder \citep{wu2023towards}. Because LLaVA-Med is not implemented with a 3D vision encoder, to ensure a fair comparison, we test it with our model's vision encoder \citep{bai2024m3d}. Similar to our method, the vision encoder is frozen and only the projection layer and LLM are trainable. To provide a comparison to our model's multi-task classifier, which is trained end-to-end with the rest of our framework, we independently train a new multi-task classifier. We use a Phi3 language model with multi-task heads to predict the multi-task outputs given the prompt and text generation. The model is trained on our train dataset and had 99.8\% accuracy on the validation set. Other hyperparameter settings mirror our method as closely as possible to ensure a fair comparison. 

\paragraph{Evaluation} For evaluating the models' task proficiency, we use accuracy for volume, shape, spread, and out-of-scope tasks, and per-label accuracy for the region task. We estimate the standard deviation using 500 bootstrap resamples.

\paragraph{Computing environment} All our experiments were mainly conducted using a single NVIDIA A100 GPU on an internal cluster. Training our model on the GLI dataset took approximately 10 hours.

\begin{table}
\setlength{\textfloatsep}{8pt plus 2pt minus 2pt}
\centering
\caption{Ablation study on the MoE architecture on the GLI validation set with accuracy metric.}
\small
\SetTblrInner{rowsep=0.95pt}
\begin{tblr}{colspec = {cccc}}
        % row{1} = {bg=gray!25},
        % cell{odd[2-6]}{1-7} = {bg=gray!10}}
 \toprule
 \makecell{Modality-level\\MoE} &
 \makecell{Token-level\\MoE} &
 \makecell{Prompt-based\\MoE weights} &
 \makecell{Task\\ Mean} \\
\midrule
 \xmark & \xmark & \xmark & 63.3 \\
 \checkmark & \xmark & \xmark & 64.1 \\
  \xmark & \checkmark &  \xmark &  64.4 \\
 \checkmark & \checkmark & \checkmark & \textbf{65.5} \\ 
\bottomrule
\end{tblr}
\label{tab:abl}
\end{table}

\begin{table}[h]
\caption{Comparison of MoE-based and shared expert approach based on task on the GLI validation set with accuracy metric}
\label{tab:moe_vs_shared_label}
\centering
\begin{tabular}{l c c c c}
\toprule
Method & Volume & Region & Shape & Spread \\
\midrule
Shared expert & 53.2 & 82.8 & 59.7 & 57.5 \\
MoE-based & \textbf{57.6} & \textbf{83.5} & \textbf{60.9} & \textbf{59.9} \\
\bottomrule
\end{tabular}
\end{table}

\begin{table}[h]
\caption{Comparison of MoE-based and shared expert approach based on finding on the GLI validation set with accuracy metric}
\label{tab:moe_vs_shared_task}
\centering
\begin{tabular}{l c c c c}
\toprule
Method & ET & SNFH & NETC & RC \\
\midrule
Shared expert & 63.2 & 70.8 & 71.3 & 48.7 \\
MoE-based & \textbf{65.1} & \textbf{72.5} & \textbf{75.1} & \textbf{49.9} \\
\bottomrule
\end{tabular}
\end{table}

\begin{table}[h]
\caption{Comparison of MoE approaches based on task on the GLI validation set with accuracy metric}
\label{tab:moe_task}
\centering
\begin{tabular}{l c c c c}
\toprule
Method & Volume & Region & Shape & Spread \\
\midrule
Modality-level MoE     & 53.7 & 83.4 & 59.7 & 59.4 \\
Token-level MoE        & 56.0 & \textbf{83.5} & 60.0 & 58.1 \\
Prompt-based MoE & \textbf{57.6} & \textbf{83.5} & \textbf{60.9} & \textbf{59.9} \\
\bottomrule
\end{tabular}
\end{table}

\begin{table}[h]
\caption{Comparison of MoE approaches based on finding on the GLI validation set with accuracy metric}
\label{tab:moe_label}
\centering
\begin{tabular}{l c c c c}
\toprule
Method & ET & SNFH & NETC & RC \\
\midrule
Modality-level MoE     & 63.5 & 71.1 & 73.9 & 48.6 \\
Token-level MoE        & 64.0 & 72.2 & 72.2 & \textbf{49.9} \\
Prompt-based MoE & \textbf{65.1} & \textbf{72.5} & \textbf{75.1} & \textbf{49.9} \\
\bottomrule
\end{tabular}
\end{table}

\begin{table}[h]
\centering
\caption{Comparison of task means with and without region information on the GLI validation set with accuracy metric.}
\begin{tabular}{l c}
\toprule
Model & Task Mean without Region Scores \\
\midrule
Without Region Information & 59.5 \\
With Region Information & \textbf{64.1} \\
\bottomrule
\end{tabular}
\label{tab:region_info}
\end{table}

\begin{table}[h]
\centering
\caption{Radiologist acceptance rate comparison between \ours and M3D.}
\begin{tabular}{l c}
\toprule
Model & Radiologist Acceptance Rate (\%) \\
\midrule
M3D & 34.1 \\
\ours & \textbf{50.0} \\
\bottomrule
\end{tabular}
\label{tab:clin_suf_val}
\end{table}

\begin{table}[h]
\centering
\caption{Comparison of model performance for differentiating primary gliomas versus secondary metastatic lesions}
\begin{tabular}{l c c}
\toprule
Model & Accuracy & AUROC \\
\midrule
M3D & 88.5 & 95.5 \\
\ours & \textbf{95.6} & \textbf{99.0} \\
\bottomrule
\end{tabular}
\label{tab:downstream_task}
\end{table}

\subsection{Results}

All model results across the evaluated datasets are presented in Table~\ref{tab:results}. Our model consistently achieves strong performance across all task categories and datasets, outperforming the second-best model by an average margin of 5.5 points (+9.1\% relative to the strongest baseline). We note a small drop in performance relative to other models on the GoAT dataset. This dataset is extremely challenging because it evaluates generalizability based on several factors with limited training data. In our experiments, we also noticed that the top three models had above a 99.8\% accuracy on out-of-scope task identification, which suggests our dataset was effective at hallucination mitigation.

\subsection{Ablation studies}
\label{sec:abl}

An ablation study on the MoE architecture is in Table~\ref{tab:abl}. Image modality-level and token-level high-level MoE experts perform better than the single projection layer baseline approach. A prompt-conditioned weighted combination of the different high-level experts performs the best.

Fine-grained results comparing our MoE-based approach and a single shared expert are in Table~\ref{tab:moe_vs_shared_label} and Table~\ref{tab:moe_vs_shared_task}. The more complex multimodal reasoning is helpful for all tasks and findings, and especially helpful for the volume task (+4.4 points and 8.3\% relative to the baseline) and NETC finding (+3.8 points and 5.3\% relative to the baseline). Fine-grained results comparing modality-level MoE, token-level MoE, as well prompt-conditioned MoE on tasks as well as findings are in Table~\ref{tab:moe_task} and Table~\ref{tab:moe_label}. Token-level MoE is stronger in the volume, region, and shape tasks while modality-level MoE is stronger in the spread task. For findings, token-level MoE is stronger with the ET, SNFH, and RC findings while modality-level MoE is stronger with the NETC finding. The prompt-conditioned MoE excels in all of them, suggesting that based on the question, the model is able to accurately combine the optimal token-level MoE or modality-level MoE blocks.

\subsection{Clinical perspective}
\label{sec:clin_utility}

From a neuroradiology standpoint, an important use case for \ours is as an adjunct to clinical expertise: producing standardized, interpretable tumor descriptors (e.g., size, morphology, extent, and coarse localization) for referring neurosurgeons. This framework also provides a pathway to more fine-grained future outputs, such as involvement with specific gyri or eloquent cortex. A unique technical challenge in neuro-oncology imaging is robustness to atypical anatomy and our results suggest the proposed approach remains effective across highly variable lesion appearances.

To quantify clinical utility under this framing, we evaluate (i) agreement between model predictions and radiologist annotations, (ii) radiologist acceptability of generated responses, and (iii) clinically relevant downstream performance. Agreement results are shown in Table~\ref{tab:clin_dataset_model_val}. We observe that \ours is comparable to radiologist-level consistency on the volume and shape tasks, indicating strong reliability for two core descriptors commonly communicated in clinical workflows. Performance is lower on region and spread, which we attribute primarily to higher noise in the synthetic labeling pipeline for these attributes and to the inherent difficulty of localization/extent estimation when anatomy is atypical or deformed. This result highlights a practical constraint for real-world deployment: model performance is bounded by label quality and by the stability of the underlying definitions.

We further probe the role of localization by appending region information to the prompt (Table~\ref{tab:region_info}), yielding an approximately 5\% increase on other task scores. This suggests (i) localization provides useful signal for estimating morphology/extent and (ii) even relatively noisy region labels can improve other clinically relevant descriptors, warranting their inclusion. This aligns with clinical practice, where coarse localization is often communicated alongside morphology/extent.

Because many descriptors are subjective, we collaborated with a radiologist to assess whether generated answers are clinically reasonable. We created a clinical validation set of 208 questions from 13 GLI test cases, focusing on volume, region, shape, and spread (questions 1--4 in Appendix~\ref{prompt_list}). We compared \ours against M3D by asking the radiologist to label each response as sufficient or lacking. \ours achieves a 50.0\% acceptability rating and improves clinical acceptability by +15.9 points (+46.6\% relative; Table~\ref{tab:clin_suf_val}). These findings support the role of \ours as a reporting adjunct for tumor descriptors.

Finally, to emphasize downstream usefulness, we consider a clinically important task: differentiating primary gliomas from secondary metastatic lesions using imaging-derived descriptors (volume, region, shape, spread). Because our models are trained with guard rails and indicate if tasks are outside of their scope, we fit a logistic regression classifier on model-generated features from the combined training set of BraTS GLI and MET. On the test set (Table~\ref{tab:downstream_task}), \ours improves over M3D by +7 accuracy points and +4 AUROC points, suggesting that the learned structured descriptors capture discriminative imaging patterns relevant to neuro-oncology workflows.

\section{Conclusion}

We present \ours, a multimodal LLM for mpMRI VQA that uses prompt-conditioned hierarchical MoE routing over modality- and token-level projection experts, enabling efficient end-to-end fine-tuning without paired image–report pretraining. We pair this model with a synthetic VQA pipeline derived from expert segmentation annotations, developed in collaboration with a radiologist, to produce standardized tumor descriptors (e.g., volume, morphology, extent, and coarse localization) that can serve as an adjunct to clinical expertise for referring neurosurgeons. Across BraTS GLI/MET/GoAT, \ours improves over strong medical VLM baselines by an average of +5.5 points and increases radiologist-rated clinical acceptability by +15.9 points, with ablations validating the contributions of modality/token experts and prompt-conditioned routing. These findings suggest that structured, interpretable outputs can remain reliable across heterogeneous lesion appearances, while also highlighting the importance of label quality for more challenging localization tasks. Future work includes extending to open-ended VQA and report generation, evaluating robustness in settings with atypical anatomy, conducting broader multi-reader validation, and performing fairness analyses.

\section{Limitations and Ethical Considerations}

This work uses publicly available, fully de-identified BraTS datasets, minimizing risks to patient privacy and data security. Our synthetic VQA are generated from segmentation annotations, and both the generated questions and model outputs underwent radiologist review to mitigate typical risks of synthetic supervision. Nonetheless, fairness and bias remain open concerns: synthetic prompts and limited demographic metadata can yield models that underperform for underrepresented groups or clinical scenarios. The model is intended for research only and must not be used for autonomous clinical decision-making; it is designed to abstain on out-of-scope queries, and any deployment would require prospective, multi-site validation under qualified clinical oversight. In future work, we will expand evaluations to demographically diverse cohorts where available, document dataset composition and known limitations, and incorporate explicit fairness analyses and bias-mitigation strategies alongside robustness and calibration assessments. 

\section{GenAI Disclosure}
\label{sec:llm_usage}
We used large language models (LLMs) to (i) improve the clarity and style of the manuscript, (ii) brainstorm refinements to the MoE-based architecture and dataset-construction procedures, (iii) draft code prototypes for selected ideas, and (iv) find potentially relevant related work. All LLM outputs were reviewed and verified by the authors before inclusion.

\bibliographystyle{ACM-Reference-Format}
\bibliography{main}
\clearpage

\appendix

\section{Additional information regarding dataset}
\label{sec:more_dataset_details}

\begin{table*}
\setlength{\textfloatsep}{8pt plus 2pt minus 2pt}
\centering
\SetTblrInner{rowsep=0.95pt}
\caption{Statistics for the synthetic VQA datasets.}
\begin{tblr}{colspec = {ccccc}}
    \toprule
    Dataset & \# questions & \# mpMRI & \# unique questions & \# unique answers \\
    \midrule
    GLI & 38,904 & 1,621 & 38,023 & 36,773 \\
    MET  & 11,718 & 651 &  11,607 & 11,284 \\
    GoAT  & 24,318 & 1,351 & 23,859 & 23,223 \\
    \bottomrule
\end{tblr}
\label{tab:gen_vqa}
\end{table*}

In the following, we will describe the formulas used to derive the shape and spread descriptors for our synthetic VQA protocol. Let $M\subset\mathbb{Z}^{3}$ be a binary mask of lesion voxels sampled with spacing $\mathbf s=(s_x,s_y,s_z)\,[\mathrm{mm}]$ (typically $s_x=s_y=s_z=1$).
Write $\Delta V = s_xs_y s_z$ for the physical volume of one voxel and $\lvert M\rvert$ for the number of lesion voxels.

\paragraph{Total volume}

  $$
    V_{\text{tot}} = \lvert M\rvert \,\Delta V\;[\mathrm{mm}^3].
  $$

\paragraph{Multiplicity} We define 26-connectivity on $M$ (scipy `ndimage.label` with a unit “ball” structuring element) and record $N_{\!c}$ and $M_1,\dots,M_{N_{\!c}}$.

\paragraph{Spread} Let the \textit{core component} index be $i^\star=\arg\max_{i}V_i$. Define

$$
  f_{\text{core}} \;=\; \frac{V_{i^\star}}{V_{\text{tot}}}\in[0,1].
$$

  $$
    \text{spread}=\begin{cases}
      \text{``single lesion''} & N_{\!c}=1,\\[2pt]
      \text{``core with satellite lesions''} & N_{\!c}>1,\;f_{\text{core}}\ge0.85,\\[2pt]
      \text{``scattered lesions''} & \text{otherwise.}
    \end{cases}
  $$

For each component $M_i$:

\paragraph{Component surface area}

Marching cubes (scikit-image `measure.marching\_cubes`) produces a triangular mesh $(\mathcal V_i,\mathcal F_i)$ in real-world coordinates. The mesh area (which we describe as the surface area) is

   $$
     A_i=\!\!\!\sum_{(p,q,r)\in\mathcal F_i}\!\!\!
           \tfrac12\bigl\|(q-p)\times(r-p)\bigr\|_2 .
   $$

\paragraph{Component volume} $V_i=\lvert M_i\rvert\,\Delta V$.

\paragraph{Component sphericity}
   $$
     \Phi_i \;=\; 
     \frac{\pi^{1/3}\,(6V_i)^{2/3}}{A_i}.
   $$

\paragraph{Component compactness}

    $$
     C_i \;=\; \frac{A_i}{V_i}.
    $$

\paragraph{Component principal-axis statistics}
Assemble voxel coordinates $\mathbf x_j=(x_j,y_j,z_j)\in\mathbb R^{3}$ for $j\in M_i$. The covariance matrix $\Sigma_i=\frac1{\lvert M_i\rvert}\sum_j(\mathbf x_j-\bar{\mathbf x}) (\mathbf x_j-\bar{\mathbf x})^{\!\top}$ yields eigenvalues $\lambda_1\ge\lambda_2\ge\lambda_3>0$.

\paragraph{Component elongation}
     $$
       E_i \;=\; \sqrt{\lambda_1/\lambda_2}.
     $$

\paragraph{Component flatness}
     $$
       F_i \;=\; \sqrt{\lambda_3/\lambda_2}.
     $$

\paragraph{Component solidity} A convex hull (scipy `ConvexHull`) provides volume $V^\mathrm{hull}_i$;

   $$
     S_i \;=\; \frac{V_i}{V^\mathrm{hull}_i}.
   $$

\paragraph{Metric aggregation}

  $$
    (\Phi,E,F,S,C)=
    \begin{cases}
      (\Phi_{i^\star},E_{i^\star},F_{i^\star},S_{i^\star},C_{i^\star}) & 
        N_{\!c}=1\text{ or }f_{\text{core}}\ge0.7,\\[4pt]
      \displaystyle\frac1{N_{\!c}}\sum_{i=1}^{N_{\!c}}
        (\Phi_i,E_i,F_i,S_i,C_i) & \text{otherwise.}
    \end{cases}
  $$

\paragraph{Shape} Convert the continuous metrics to one of five categories:

   $$
     \text{shape}=\begin{cases}
     \text{``focus''} & V_{\text{tot}}<0.1\;\text{cm}^3
     \quad(\;V_{\text{tot}}\times10^{-3}<0.1\;) \\[2pt]
       \text{``round''} & \Phi\ge0.80\;\land\;E\le1.4,\\[2pt]
       \text{``oval''} & 0.50\le\Phi<0.80\;\land\;1.4<E\le3.0,\\[2pt]
       \text{``elongated''} & E>3.0,\\[2pt]
       \text{``irregular''} & \text{otherwise.}
     \end{cases}
   $$

The thresholds were set empirically on 10 annotated mpMRIs from BraTS-GLI for validation set and match radiologists’ qualitative intuition of near-spherical, mildly flattened, and strongly stretched geometries. All computations are implemented in Python using scipy, scikit-image, numpy, and ndimage as shown in the listing above.

\paragraph{Question augmentation details} We use ChatGPT to generate question augmentations of our multitask dataset. For generating question augmentations for the standard multi-task prompts, we first provide this prompt ``Please produce hundred alternative wordings that a radiologist may use for the following question and answer. Please include everything surrounded by curly braces \{\} as they are because they are placeholders. Please generate the reworded question starting with ``Q:'' and reworded answer starting with ``A:'' and separate each generated question-answer pair with a newline. Please do not produce any additional text.'' and append this to each of the multitask prompts below. We produce 40 repetitions with a temperature of $1.0$, top p of $1$, and model ``gpt-4o-mini-2024-07-18''. 

\begin{enumerate}
\label{prompt_list}
    \item Q: How large is the volume covered by \{label\}? A: The overall volume of \{label\} is \{volume\}. 
    \item Q: Which region(s) of the brain is \{label\} located in? A: The \{label\} is located in \{regions\}. 
    \item Q: What is the shape of \{label\}? A: The shape of \{label\} is \{shape\}. 
    \item Q: How spread out is \{label\}? A: The spread of \{label\} is \{spread\}.
    \item Q: How large is the volume of \{label\} and where is it located? A: The overall volume of \{label\} is \{volume\}, and it is located in \{regions\}.
    \item Q: How large is the volume of \{label\} and what is its shape? A: The overall volume of \{label\} is \{volume\}, and its shape is described as \{shape\}.
    \item Q: How large is the volume of \{label\} and how spread out is it? A: The overall volume of \{label\} is \{volume\}, and it is characterized as \{spread\}. 
    \item Q: In which region is \{label\} and what is its shape? A: The \{label\} is located in \{regions\}, and its shape is described as \{shape\}. 
    \item Q: In which region is \{label\} and how spread out is it? A: The \{label\} is located in \{regions\}, and it is characterized as \{spread\}.
    \item Q: What is the shape of \{label\} and how spread out is it? A: The shape of \{label\} is described as \{shape\}, and it is characterized as \{spread\}.
    \item Q: What is the volume, region, and shape of \{label\}? A: The overall volume of \{label\} is \{volume\}, it is located in \{regions\}, and its shape is described as \{shape\}. 
    \item Q: What is the volume, region, and spread of \{label\}? A: The overall volume of \{label\} is \{volume\}, it is located in \{regions\}, and it is characterized as \{spread\}. 
    \item Q: What is the volume, shape, and spread of \{label\}? A: The overall volume of \{label\} is \{volume\}, its shape is described as \{shape\}, and it is characterized as \{spread\}. 
    \item Q: What is the region, shape, and spread of \{label\}? A: The \{label\} is located in \{regions\}, its shape is described as \{shape\}, and it is characterized as \{spread\}. 
    \item Q: What is the volume, region, shape, and spread of \{label\}? A: The overall volume of \{label\} is \{volume\}, it is located in \{regions\}, its shape is described as \{shape\}, and it is characterized as \{spread\}. 
\end{enumerate}

For generating question augmentations for the partially out-of-scope multi-task prompts, we first provide this prompt ``Please produce hundred alternative wordings that a radiologist may use for the following question and answer and incorporate an additional clinical task or tasks which the model cannot solve in the reworded question. These can be before, after, or interspersed between the other tasks (please make sure to vary the order and number of out-of-scope tasks). Do not mention that the model cannot answer these in the question; however, indicate that the model cannot answer that part of the question in the reworded answer (potentially using different phrasings). The model can describe the volume, brain region, shape, and spread of \{label\} which is the region of interest. Please include everything surrounded by curly braces \{\} as they are because they are placeholders. Please generate the reworded question starting with ``Q:'' and reworded answer starting with ``A:'' and do not produce any additional text.'' and append this to each of the multitask prompts above. We produce 10 repetitions with a temperature of $1.0$, top p of $1$, and model ``gpt-4o-2024-08-06''.

For generating question augmentations for completely out-of-scope prompts, we first provide this prompt ``Please produce a hundred questions (with one or more tasks) that a radiologist may ask that the model does not have information to answer. The model can describe the volume, brain region, shape, and spread of \{label\} which is the region of interest. Please include \{label\} in the question but do not include anything else with curly braces. In the answer, please indicate the model cannot answer the question (potentially using different phrasings). Please generate the question starting with ``Q:'' and answer starting with ``A:'' and do not produce any additional text.'' We produce 10 repetitions with a temperature of $1.0$, top p of $1$, and model ``gpt-4o-mini-2024-07-18''.

After generating the question augmentations, we apply a rules-based quality check: questions and answers are properly split with "Q:" and "A:" respectively, contents within the curly braces are retained for easy formatting with Python, and that the responses are in English. Then, for each finding and mpMRI in each dataset, we sample four multitask questions without replacement such that each major task is addressed in at least one question, one partially out-of-scope question, and one completely out-of-scope question. Examples of generated question types can be seen in Table~\ref{tab:qual_vqa}. Additionally, all the generated test set question candidates pass an automated data cleaning step using ChatGPT (model ``gpt-4o-2024-08-06'') to ensure high quality evaluation. See the prompts below for the three different question types.

\begin{enumerate}
\label{data_cleaning_prompt_list}
    \item Determine whether the reworded question is a faithful, clinically appropriate rephrasing of the original question. Please ignore {valid} as this is a placeholder. Output VALID or INVALID with no additional text. 
    \item Determine whether the reworded question is a faithful, clinically appropriate rephrasing of the original question. The reworded question also has an additional query not in the original question. Please also determine whether the added query is a valid medical question. Please ignore {label} as this is a placeholder. Output VALID or INVALID with no additional text.
    Example 1:
    Original Q: How large is the volume of {label} and what is its shape?
    Reworded Q: What is the mass and form of {label}, and engage in patient counseling?
    INVALID
    Example 2:
    Original Q: How large is the volume of {label} and what is its shape? 
    Reworded Q: Can you estimate the volume and layout of {label}, and administer a flu vaccine?
    VALID
    \item Determine whether the reworded question is a valid medical question. Please ignore {valid} as this is a placeholder. Output VALID or INVALID with no additional text.
\end{enumerate}

To assess the quality of our synthetic questions, a radiologist evaluated the clarity of 160 synthetic questions from 10 mpMRIs from the BraTS-GLI test set using a binary scoring system (1 = valid, 0 = invalid). Prior to the LLM automated data cleaning, the synthetic questions achieved a 92.2\% validity rate. After automated data cleaning, the synthetic questions achieved a 95.9\% validity rate.

After the application of our synthetic VQA protocol, the percentage frequency of each task per question for all the generated datasets can be seen in Table~\ref{tab:gli_data_results}, Table~\ref{tab:met_data_results}, and Table~\ref{tab:goat_data_results}.

\begin{table}
\setlength{\textfloatsep}{8pt plus 2pt minus 2pt}
\centering
\caption{Qualitative examples illustrating the four question types generated by our pipeline.}
\begin{tblr}{
  colspec = {Q[2.3] Q[6] Q[6]}, % Q is the Tabularray auto‑wrap column
}
Type & Question & Answer \\
\midrule
Single‑task &
Could you identify the shape of the Resection Cavity? &
The shape of the Resection Cavity is \textbf{irregular}. \\
Multi‑task &
How do you quantify the volume of Surrounding Non‑enhancing FLAIR hyperintensity, and what does its distribution look like? &

The total volume of Surrounding Non-enhancing FLAIR hyperintensity is described as \textbf{1–5\%}, characterized as \textbf{core with satellite lesions}. \\
Partial‑out-of-scope &
What areas does Surrounding Non‑enhancing FLAIR hyperintensity encompass, what is its volume, and synthesize findings into a comprehensive care plan? &
Surrounding Non-enhancing FLAIR hyperintensity has a volume of \textbf{1–5\%} in \textbf{cerebellum, frontal and parietal}. Synthesizing this into a care plan is \textit{outside my domain}. \\
Out-of-scope &
How do genetic factors influence the development of Non‑Enhancing Tumor in adolescents? &
\textit{I cannot provide information} regarding the influence of genetic factors on the development of Non-Enhancing Tumor in adolescents. \\
\end{tblr}
\label{tab:qual_vqa}
\end{table}

\begin{table}[h]
\setlength{\textfloatsep}{8pt plus 2pt minus 2pt}
\centering
\caption{Percentage frequency of each task label per question for the GLI dataset}
\SetTblrInner{rowsep=0.95pt}
\begin{tblr}{colspec = {ccc}}
    \toprule
\makecell{Task} & \makecell{Label name} & \makecell{Label frequency} \\
\midrule
\SetCell[r=6]{c}{Volume}
& Unspecified & 52.4 \\
& N/A & 12.3 \\ 
& $<$1\% & 12.7 \\
& 1-5\% & 13.1 \\
& 5-10\% & 5.2 \\
& 10-25\% & 3.8 \\
& 25-50\% & 0.5 \\
& 50-75\% & 0.0 \\
\midrule
\SetCell[r=11]{c}{Region}
& Unspecified & 53.2 \\
& N/A & 12.2 \\ 
& frontal & 23.8 \\
& parietal & 20.8 \\
& occipital & 13.2 \\
& temporal & 17.2 \\
& limbic & 21.7 \\
& insula & 14.7 \\
& subcortical & 14.9 \\
& cerebellum & 2.9 \\
\midrule
\SetCell[r=7]{c}{Shape}
& Unspecified & 52.4 \\
& N/A & 12.3 \\ 
& focus & 1.0 \\
& round & 4.7 \\
& oval & 6.9 \\
& elongated & 0.4 \\
& irregular & 22.4 \\
\midrule
\SetCell[r=5]{c}{Spread}
& Unspecified & 53.4 \\
& N/A & 12.2 \\ 
& single lesion & 6.9 \\
& core with satellite lesions & 20.9 \\
& scattered lesions & 6.5 \\
\midrule
\SetCell[r=2]{c}{Out-of-scope}
& Not out-of-scope & 66.7 \\
& Out-of-scope & 33.3 \\ 
\midrule
\end{tblr}
\label{tab:gli_data_results}
\end{table}

\begin{table}[h]
\setlength{\textfloatsep}{8pt plus 2pt minus 2pt}
\centering
\caption{Percentage frequency of each task label per question for the MET dataset}
\SetTblrInner{rowsep=0.95pt}
\begin{tblr}{colspec = {ccc}}
    \toprule
\makecell{Task} & \makecell{Label name} & \makecell{Label frequency} \\
\midrule
\SetCell[r=6]{c}{Volume}
& Unspecified & 51.8 \\
& N/A & 8.5 \\ 
& $<$1\% & 24.7 \\
& 1-5\% & 9.2 \\
& 5-10\% & 2.7 \\
& 10-25\% & 2.7 \\
& 25-50\% & 0.4 \\
& 50-75\% & 0.0 \\
\midrule
\SetCell[r=11]{c}{Region}
& Unspecified & 53.0 \\
& N/A & 17.1 \\ 
& frontal & 19.5 \\
& parietal & 16.1 \\
& occipital & 14.5 \\
& temporal & 14.5 \\
& limbic & 9.2 \\
& insula & 6.3 \\
& subcortical & 7.3 \\
& cerebellum & 12.8 \\
& brainstem & 4.1 \\
\midrule
\SetCell[r=7]{c}{Shape}
& Unspecified & 51.4 \\
& N/A & 8.3 \\ 
& focus &  2.8 \\
& round & 13.5 \\
& oval & 4.3 \\
& elongated & 0.2 \\
& irregular & 19.5 \\
\midrule
\SetCell[r=5]{c}{Spread}
& Unspecified & 53.1 \\
& N/A & 7.9 \\ 
& single lesion & 7.4 \\
& core with satellite lesions & 12.9 \\
& scattered lesions & 18.7 \\
\midrule
\SetCell[r=2]{c}{Out-of-scope}
& Not out-of-scope & 66.7 \\
& Out-of-scope & 33.3 \\ 
\midrule
\end{tblr}
\label{tab:met_data_results}
\end{table}

\begin{table}[h]
\setlength{\textfloatsep}{8pt plus 2pt minus 2pt}
\centering
\caption{Percentage frequency of each task label per question for the GoAT dataset}
\SetTblrInner{rowsep=0.95pt}
\begin{tblr}{colspec = {ccc}}
    \toprule
\makecell{Task} & \makecell{Label name} & \makecell{Label frequency} \\
\midrule
\SetCell[r=6]{c}{Volume}
& Unspecified & 52.2 \\
& N/A & 2.2 \\ 
& $<$1\% & 10.3 \\
& 1-5\% & 18.6 \\
& 5-10\% & 8.9 \\
& 10-25\% & 7.2 \\
& 25-50\% & 0.6 \\
\midrule
\SetCell[r=11]{c}{Region}
& Unspecified & 52.9 \\
& N/A & 2.3 \\ 
& frontal & 30.0 \\
& parietal & 23.4 \\
& occipital & 17.5 \\
& temporal & 29.7 \\
& limbic & 29.8 \\
& insula & 25.9 \\
& subcortical & 27.9 \\
& cerebellum & 9.8 \\
& brainstem & 8.0 \\
\midrule
\SetCell[r=7]{c}{Shape}
& Unspecified & 51.9 \\
& N/A & 2.1 \\ 
& focus & 0.5 \\
& round & 6.1 \\
& oval & 3.2 \\
& elongated & 0.1 \\
& irregular & 36.0 \\
\midrule
\SetCell[r=5]{c}{Spread}
& Unspecified & 53.3 \\
& N/A & 1.9 \\ 
& single lesion & 5.3 \\
& core with satellite lesions & 32.7 \\
& scattered lesions & 6.7 \\
\midrule
\SetCell[r=2]{c}{Out-of-scope}
& Not out-of-scope & 66.7 \\
& Out-of-scope & 33.3 \\ 
\midrule
\end{tblr}
\label{tab:goat_data_results}
\end{table}

\section{Additional Ablation results}
\label{sec:addl_ablation}
Additional ablation results validating the number of high-level experts, softmax versus sigmoid for summing lower-level experts, and concatenation versus element-wise summing of vision tokens are in Table~\ref{tab:num_experts}, Table~\ref{tab:sigmoid_vs_softmax}, and Table~\ref{tab:proj_fusion} respectively. In Table~\ref{tab:loss}, we see a comparison of our model performance trained with our multi-task loss versus the next-token prediction baseline loss. There is a significant performance improvement with our multi-task loss.

To qualitatively evaluate our architecture, we construct all 60 template task prompts from our GLI dataset (four findings $\times$ 15 task combinations = 60 template prompts) and input them into our model's high-level router to generate high-level expert weight vectors. We then calculate the correlation between these weight vectors and generate a heatmap, which is in Figure~\ref{fig:heatmap}. There's high correlation between expert weight vectors within the same finding, suggesting similar image features are extracted. For findings that are closer anatomically, such as non-enhancing tumor core and enhancing tissue, there is also relatively high correlation between the expert weight vectors, which suggests similar extracted image features. For findings like resection cavity and surrounding FLAIR hyperintensity that are more diverse anatomically from the other findings, there's much lower correlation, which suggests more dissimilar extracted features.

\section{Reproducibility statement}

We use a publicly available dataset and detail the full data-generation pipeline in Sections~\ref{sec:auto_vqa} and~\ref{sec:dataset_details}, with additional information in Appendix~\ref{sec:more_dataset_details}. We provide the github link our model code in the abstract, and report all experimental settings and computational resources in Section~\ref{sec:exp_settings}.

\begin{table*}
\setlength{\textfloatsep}{8pt plus 2pt minus 2pt}
\centering
\caption{Inter-annotator agreement (accuracy) between synthetic data and radiologists}
\SetTblrInner{rowsep=0.95pt}
\begin{tblr}{colspec = {ccccc}}
    \toprule
\makecell{Agreement between synthetic data and} & \makecell{Volume} & \makecell{Region} & \makecell{Shape} & \makecell{Spread} \\
\midrule 
Radiologist 1 & 51.0 & 74.2 & 56.0 & 39.0 \\
Radiologist 2 & 49.0 & 75.7 & 43.0 & 48.0 \\
\midrule
\end{tblr}
\label{tab:clin_dataset_val}
\end{table*}

\begin{table*}
\setlength{\textfloatsep}{8pt plus 2pt minus 2pt}
\centering
\caption{Inter-annotator agreement (accuracy) between model predictions and radiologists and between both radiologists}
\SetTblrInner{rowsep=0.95pt}
\begin{tblr}{colspec = {ccccc}}
    \toprule
\makecell{Agreement between} & \makecell{Volume} & \makecell{Region} & \makecell{Shape} & \makecell{Spread} \\
\midrule 
Model and Radiologist 1 & 43.0 & 70.9 & 47.0 & 23.0 \\
Model and Radiologist 2 & 43.0 & 71.2 & 49.0 & 38.0 \\
Both Radiologists & 38.0 & 85.5 & 50.0 & 62.0 \\
\midrule
\end{tblr}
\label{tab:clin_dataset_model_val}
\end{table*}

%\begin{table*}
%\setlength{\textfloatsep}{8pt plus 2pt minus 2pt}
%\centering
%\caption{Radiologist acceptance rate comparison by task between \ours and M3D}
%\SetTblrInner{rowsep=0.95pt}
%\begin{tblr}{colspec = {cccccc}}
%    \toprule
%\makecell{Model} & \makecell{Volume} & \makecell{Region} & \makecell{Shape} & \makecell{Spread} & \makecell{Mean} \\
%\midrule 
%\ours & 65.4 & 28.8 & 67.3 & 38.4 & 50.0 \\
%M3D & 26.9 & 21.2 & 61.5 & 26.9 & 34.1 \\
%\midrule
%\end{tblr}
%\label{tab:rad_accpt_by_task}
%\end{table*}

\begin{table}[h]
\setlength{\textfloatsep}{8pt plus 2pt minus 2pt}
\centering
\caption{Model performance comparison with different number of high-level experts on the GLI validation set with accuracy metric.}
\SetTblrInner{rowsep=0.95pt, colsep=3.5pt}
\begin{tblr}{colspec = {lc}}
 \toprule
 \makecell{Number of blocks} & \makecell{Task Mean}\\
\midrule
12 & 64.5 \\
16 & \textbf{65.5} \\
20 & 65.0 \\
\bottomrule
\end{tblr}
\label{tab:num_experts}
\end{table}

\begin{table}[h]
\setlength{\textfloatsep}{8pt plus 2pt minus 2pt}
\centering
\caption{Model performance with softmax versus sigmoid for summing lower-level experts on the GLI validation set with accuracy metric.}
\SetTblrInner{rowsep=0.95pt, colsep=3.5pt}
\begin{tblr}{colspec = {lc}}
 \toprule
 \makecell{Method} & \makecell{Task Mean}\\
\midrule
sigmoid & 64.8 \\
softmax &  \textbf{65.5} \\
\bottomrule
\end{tblr}
\label{tab:sigmoid_vs_softmax}
\end{table}

\begin{table}[h]
\caption{Comparison of projection and fusion methods on the GLI validation set with accuracy metric.}
\centering
\begin{tabular}{l l c}
\toprule
Projection Method & Fusion Method & Task Mean \\
\midrule
Shared expert & concatenation & 52.4 \\
Shared expert & sum & 51.1 \\
Shared expert & learned weighted sum & 63.3 \\
MoE-based & sum & \textbf{65.5} \\
\bottomrule
\end{tabular}
\label{tab:proj_fusion}
\end{table}

\begin{table}[t]
\setlength{\textfloatsep}{8pt plus 2pt minus 2pt}
\centering
\caption{Model comparison with multi-task loss on the GLI validation set with accuracy metric.}
\SetTblrInner{rowsep=0.95pt, colsep=3.5pt}
\begin{tblr}{colspec = {lc}}
 \toprule
 \makecell{Method} & \makecell{Task Mean}\\
\midrule
\ours without multi-task loss & 59.5 \\
\ours with multi-task loss & \textbf{65.5} \\
\bottomrule
\end{tblr}
\label{tab:loss}
\end{table}

\begingroup
\setlength{\textfloatsep}{8pt plus 2pt minus 2pt} % space below top floats
\begin{figure*}[t]
    \centering
    \includegraphics[width=\linewidth,trim=6pt 6pt 6pt 6pt]{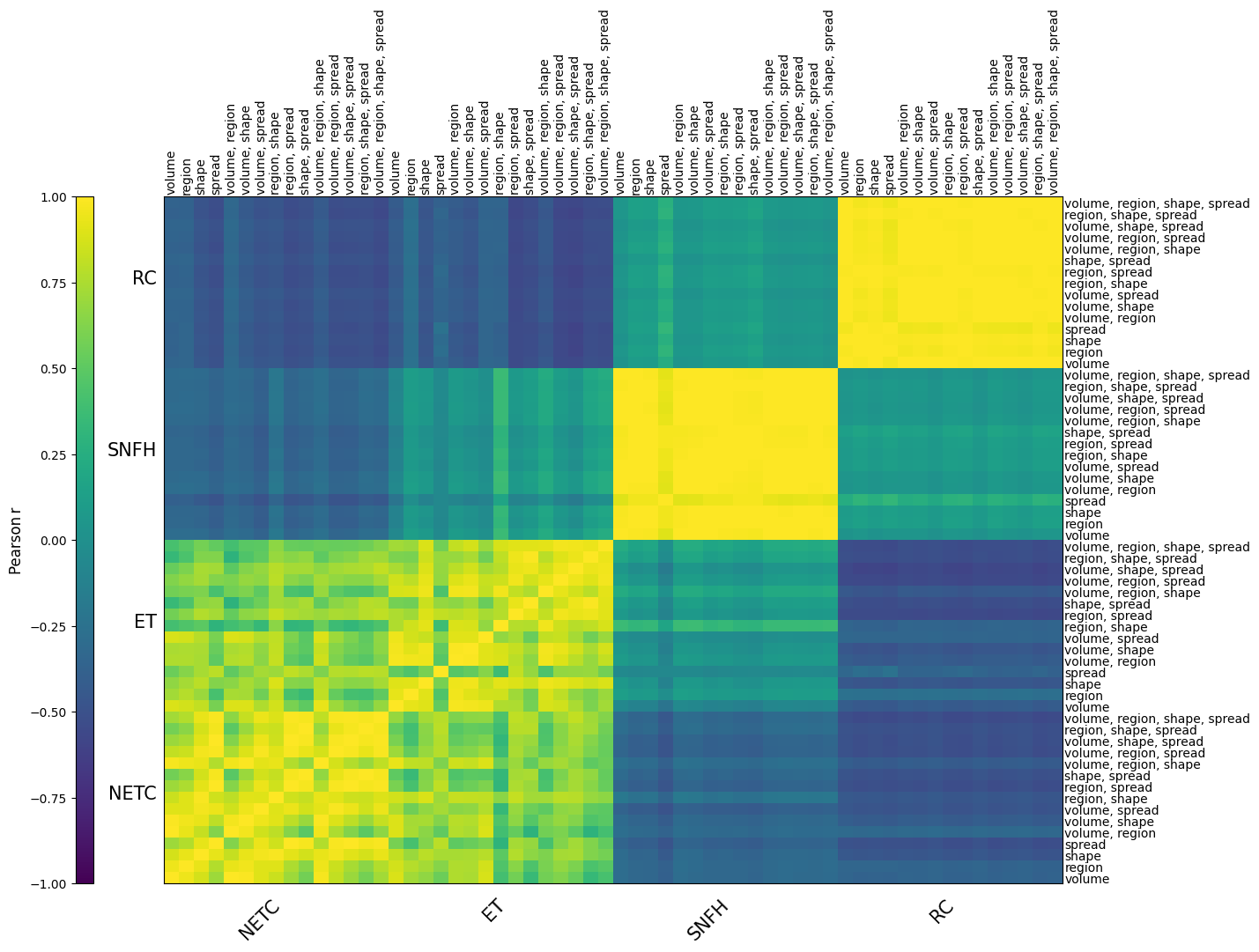}
    \caption{Heatmap for correlation between high-level expert weight vectors for standard prompts in the GLI dataset. NETC = non-enhancing tumor core, ET = enhancing tissue, SNFH = surrounding FLAIR hyperintensity, RC = resection cavity.}
    \Description{Heatmap for correlation between high-level expert weight vectors for standard prompts in the GLI dataset. NETC = non-enhancing tumor core, ET = enhancing tissue, SNFH = surrounding FLAIR hyperintensity, RC = resection cavity.}
    \label{fig:heatmap}
\end{figure*}
\endgroup

\clearpage

\end{document}